\pdfoutput=1

\documentclass[11pt]{article}

\usepackage[]{acl}

\usepackage{multicol}
\usepackage{amsmath}
\usepackage{times}
\usepackage{latexsym}
\usepackage{float}
\usepackage{tabularx}
\usepackage{tikz}
\usepackage{pgfplots}

\usepackage{placeins}

\usepackage[T1]{fontenc}

\usepackage[utf8]{inputenc}

\usepackage{microtype}
\usepackage{multirow}
\usepackage{multicol}

\usepackage{inconsolata}
\usepackage{tabularx}
\usepackage{colortbl}

\usepackage{graphicx}

\usepackage{booktabs}
\usepackage{authblk}

\usepackage{graphicx}

\title{Examining Spanish Counseling with MIDAS: \\ a Motivational Interviewing Dataset in Spanish}

\author{\textbf{Aylin Gunal}\thanks{Equal contribution.}$^{\dagger}$ \hspace{0.3cm} \textbf{Bowen Yi}${^*}$$^{\dagger}$\hspace{0.3cm} \textbf{John Piette}$^{\dagger}$ \hspace{0.5cm}
        \textbf{Rada Mihalcea}$^{\dagger}$ \hspace{0.3cm} \textbf{Verónica Pérez-Rosas$^{\ddagger}$}\\
       $^{\dagger}$ University of Michigan, Ann Arbor \\
        $^{\ddagger}$Texas State University, San Marcos \\
        \texttt{\{gunala, bowenyi, jpiette, mihalcea\}@umich.edu}, \texttt{vperezr@txstate.edu}}

\begin{document}

\maketitle
\begin{sloppypar}
\begin{abstract}


Cultural and language factors significantly influence counseling, but Natural Language Processing research has not yet examined whether the findings of conversational analysis for counseling conducted in English apply to other languages. This paper presents a first step towards this direction. We introduce MIDAS (Motivational Interviewing Dataset in Spanish), a counseling dataset created from public video sources that contains expert annotations for counseling reflections and questions. 
Using this dataset, we explore language-based differences in counselor behavior in English and Spanish and develop classifiers in monolingual and multilingual settings, demonstrating its applications in counselor behavioral coding tasks.
\end{abstract}

\section{Introduction}

A growing number of natural language processing (NLP) research studies focus on mental and behavioral health issues, covering applications such as building automated chatbots to simulate counselors~ \cite{li2024mentalarenaselfplaytraininglanguage, chiu2024computationalframeworkbehavioralassessment, qiu2024interactiveagentssimulatingcounselorclient, info:doi/10.2196/52500}, monitoring patients' mental states ~\cite{chancellor2020methods, nie2024llmbasedconversationalaitherapist}, or building feedback systems to aid counselor training~\cite{sharma2023human, shen-etal-2020-counseling, li2024automatic, shen-etal-2022-knowledge}. Although this body of work seeks to address the growing need for mental health support around the world, the majority  of it has only focused on English. 
This can be partially attributed to the lack of counseling datasets in other languages, which are difficult to obtain due to the private nature of counseling interactions and the need for expert annotations. 

Patients seeking mental health care struggle to find adequate resources, especially when they are not native speakers \cite{access-to-mh}.
Studies in clinical psychotherapy have shown that cultural differences between patients and providers can lead to disparities in quality of mental health care due to unsuccessful interactions \cite{oh2016culture}. 
This highlights the importance of collecting and using culturally diverse counseling datasets when developing NLP-based tools that support counseling practice.  

In this study, we introduce MIDAS (\textbf{M}otivational \textbf{I}nterviewing \textbf{Da}taset in \textbf{S}panish), a new dataset of Spanish counseling conversations conducted using Motivational Interviewing (MI), a counseling style that focuses on eliciting patients' motivation to change \cite{miller2012motivational}. 
We use MIDAS to explore the differences in conversational strategies used by Spanish and English MI counselors. We also conduct classification experiments to classify counselor behaviors using monolingual and multilingual models. Our results show that models trained on Spanish data outperform those trained on English, highlighting the need for language-specific datasets in psychotherapy research.

\section{Related Work}

The language used in counseling varies based on the demographic and cultural background of both counselors and patients \cite{loveys-etal-2018-cross, guda-etal-2021-empathbert}, underscoring the importance of considering diversity in user identities when designing NLP systems for mental health. 


Despite growing interest in developing NLP methods for understanding counseling conversations, very few non-English datasets are publicly available, further limiting NLP research in multilingual mental healthcare. GlobHCD
\cite{meyer-elsweiler-2022-glohbcd}
is a German dataset with naturalistic interactions around changing health behavior. The interactions were obtained from participants in an online mental health forum and annotated with MI labels. Although the code to replicate the dataset is available, the annotated dataset is not publicly available.
BiMISC is a Dutch dataset that contains bilingual MI conversations manually annotated with counselor and client behaviors~\cite{dutch-mi}. Similarly,  \citet{hebrew-counseling} collected a dataset of real conversations between patients and mental health counselors and annotated the conversations with behavioral codes based on the contribution of the speaker.

The broader landscape of mental health applications for non-English NLP contains a larger body of work. Social media and text communication platforms are popular avenues for sourcing data.
The Chinese PsyQA dataset contains annotated question-answer pairs from an online mental health service \cite{chinese-mi}. The HING-POEM dataset in Hinglish examines politeness in mental health and legal counseling conversations \cite{hinglish-mi}, and research on interactions in Kenyan WhatsApp groups for peer support studies sentiment among youth living with HIV \cite{sheng-mi}. Additionally, previous work  has sourced data from social media for mental illness prediction \cite{prieto2014twitter, lopez-ubeda-etal-2019-detecting}. 
An alternative to direct data collection  is to use machine translation from high-resource to low-resource languages ~\cite{arabic-medical-llm, polish-mh-dialogues}, but this comes with the potential cost of cultural information loss.


Our study introduces the first Spanish MI dataset, filling a critical gap in the literature and offering a valuable resource for NLP researchers working on mental health applications.

\section{Motivational Interviewing Dataset in Spanish (MIDAS)} \label{section-data-collection}

\subsection{Data Collection} \label{subsection-data-collection}

We manually collect video recordings of MI interactions in Spanish from YouTube, an online video platform. We conduct keyword-based searches in Spanish for: 
\textit{entrevista motivacional } (motivational interviewing), 
\textit{demostración de entrevista motivacional} (demonstration of motivational interviewing), 
\textit{simulación de entrevista motivacional } (simulation of motivational interviewing), 
\textit{entrevista motivacional juego de roles} (motivational interviewing role playing) 
and \textit{entrevista motivacional en español} (motivational interview in Spanish). 
We select videos in Spanish, mentioning MI as the primary counseling strategy, having only two participants (i.e., counselor and patient), addressing a behavior change (e.g., smoking cessation), and containing minimal interruptions. 

The final set includes 74 Spanish counseling conversations by Spanish speakers from various geographic locations, including Spanish-speaking countries in Latin America as well as Spain. Conversations show Spanish MI demonstrations by professional counselors and MI role-play counseling by psychology students and discuss various behavioral health topics such as alcohol consumption, substance abuse, stress management, and diabetes management. 

\begin{table}[t]
\centering
\small
\resizebox{\columnwidth}{!}{%
\begin{tabular}{l|cccccc} \hline
{\multirow{2}{*}{Speaker}} &\multicolumn{2}{c}{Words}	&\multicolumn{2}{c}{Turns}   &\multicolumn{2}{c}{Words/turn}\\ \cline{2-7}
                    &Avg       &SD	                &Avg       &SD                    &Avg       &SD    \\ \hline

Counselor    &  673.52   & 589.44   &  20.35   &  14.64   &  33.09   &  40.96 \\
Client    &  501.67  &  382.09  &  19.83   &  14.41   &  25.28   &  30.33 \\ 
All     & 1190.77 & 919.36 & 40.78  & 29.31  & 29.19 & 36.27 \\ \hline
    \end{tabular}
}
    \caption{Word-level and turn-level statistics for the MIDAS dataset. }
    \label{tab:word-turn-stats-sp}

\end{table}
\textbf{Preprocessing and Transcription.} We preprocess the videos to remove introductory remarks and narratives. We then automatically transcribe and diarize the videos using Amazon Transcription\footnote{\url{https://aws.amazon.com/transcribe/}} services.  Next, we manually label the conversation participants as either a counselor or a client. Finally, the transcriptions are manually reviewed by two native Spanish speakers. 
Word-level and turn-level statistics of the final transcription set are provided in Table~\ref{tab:word-turn-stats-sp}.

\begin{table*}[t]
    \centering
    \small
    \begin{tabularx}{\textwidth}{lXc} 
         \hline
          & Transcript & Code\\ \hline
T &  \texttt{En estos años desde que le diagnosticaron diabetes ¿ha realizado algún cambio en su alimentación ? 
Quisiera comenzar tal vez a cambiar su manera de comer? ¿Qué cosas cree usted que pudiera ser capaz de hacer? ¿Con que le gustaría empezar?} &{\sc Quest} \\
& In these years since you were diagnosed with diabetes, have you made any changes to your diet? Would you like to perhaps start changing the way you eat? What things do you think you might be able to do? What would you like to start with?  & \\ 

C &   \texttt{Este... pues,  en lo especial a mi me gusta mucho ir a la panadería  ... podría limitar eso una vez a la semana}  & \\
& Um... well, specifically, I really enjoy going to the bakery ... I could limit that to once a week. & \\
T &  \texttt{Claro, podemos empezar dejando eso, el pan primero.  También podría sugerir otras ideas más adelante, si usted se siente cómoda. Tal vez a cambiar un poco, no se incluye un poco de ejercicio en su estilo de vida. Podríamos llegar a dejar algo más aparte del pan, si usted se siente cómoda al respecto.}  & {\sc REF  }\\
& Sure, we can start by cutting that out the bread first. I could also suggest other ideas later if you feel comfortable with it. Maybe little changes, I am not sure if you include exercise in your lifestyle. We could reduce something else besides the bread, if you feel comfortable with that. & \\
 \hline
    \end{tabularx}
    \caption{Transcript excerpt from an Spanish MI session between therapist (T) and client (C). MI codes include Reflection (REF) and Question (QUEST). 
    }
    \label{tab:codingSample}
\end{table*} 



         



\subsection{Annotation of Counselor Behavior}

We annotate the dataset for counselor questions and reflections, two counseling skills often studied in previous work~\cite{perez-rosas-etal-2019-makes, welivita-pu-2022-curating}. We use ITEM\footnote{\url{https://es.motivationalinterviewing.org/motivational-interviewing-resources}} (Integridad del Tratamiento de la Entrevista Motivacional), the Spanish version of the Motivational Interviewing Treatment Integrity (MITI) \cite{moyers2003motivational} coding scheme, the current gold standard for evaluating MI proficiency.

We recruit and pay three Spanish-speaking counselors with MI experience to annotate the conversations. Two are native speakers and the third speaks Spanish as a second language.  Before annotation, we evaluated interannotator reliability in five conversations, achieving a 92\% intraclass correlation
for reflections and questions, indicating good level of agreement. 
Annotation is conducted by selecting text spans for counselor turns in the transcript using Taguette,\footnote{\url{/www.taguette.org/}} a qualitative annotation platform. The final annotation set consists of 884 questions and 415 reflections. 
An annotated transcript excerpt from our dataset is shown in Table~\ref{tab:codingSample}.

\section{Analyzing Conversational Strategies of Spanish-Speaking Counselors}

We explore culture-specific strategies that Spanish-speaking counselors use in MI-style counseling by conducting language-based comparisons against MI counseling in English. We focus on conversational aspects previously identified as relevant for counseling quality, such as conversational dynamics, language use, and sentiment expressed during conversations~\cite{althoff-etal-2016-large, perez-rosas-etal-2019-makes}. 



During our analyses, we use an English counseling dataset~\cite{perez-rosas-etal-2018-analyzing} compiled with the same methodology as our Spanish dataset. It includes labels for counselor quality (low and high), as well as annotations for questions and reflections. Our analysis uses the 72 high-quality sessions available in the dataset. 
On an important note, although our dataset lacks evaluations of counseling proficiency, we assume that counselors exhibit desirable behaviors during conversations, designed to show MI skills. We instead use the reflection-to-question ratio (R:Q) as a proficiency indicator~\cite{moyers2016motivational}. 
The resulting small difference between the average ratios (0.59 for Spanish,  0.64 for English) suggests that the Spanish MI counselors in MIDAS have  proficiency levels in MI similar to the counselors represented in the English dataset.

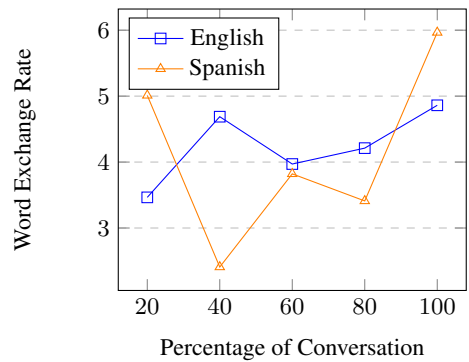
\begin{figure}[!t]
\centering
\small
\begin{tikzpicture}
\begin{axis}[
    width=.8\columnwidth,
    xlabel={Percentage of Conversation},
    ylabel={Word Exchange Rate},
    xtick={20,40,60,80,100},
    ytick={2, 3, 4, 5, 6},
    legend pos=north west,
    ymajorgrids=true,
    grid style=dashed,
]

\addplot[
    color=blue,
    mark=square,
    ]
    coordinates {
    (20,3.464642667225855)
    (40,4.687988709000965)
    (60,3.9701977312366576)
    (80,4.2114241957197915)
    (100,4.860977398525937)
    };
    \addlegendentry{English}
    \addplot[
    color=orange,
    mark=triangle,
    ]
    coordinates {
    (20,5.011056181021869)
    (40,2.4085142391127423)
    (60,3.821182476393564)
    (80,3.410366654243342)
    (100,5.965185643313296)
    
    };
    \addlegendentry{Spanish}

\end{axis}
\end{tikzpicture}
\caption{Mean word exchange rates across Spanish and English conversations.} \label{fig:exchange-rate}
\end{figure}

\noindent
\textbf{Conversation Word Exchange.} We analyze the average word exchange between counselors and clients in English and Spanish. The exchange rate is the ratio of words spoken by counselors to clients. Figure \ref{fig:exchange-rate} indicates that the Spanish exchange rate varies more over the duration of a conversation, suggesting that Spanish MI counselors speak more than their clients. 
In contrast, the exchange rate for English conversations increases slightly over the session. These differences could point to the conversational dynamics shown in clinical interactions in Spanish-speaking communities, where care providers seem to hold the higher ground during clinical conversations \citep{thompson2022cultural,coulter2003european,GimnezMoreno2022TheEO}.   



\begin{table*}[t]
    \centering
    \small
\begin{tabular}{lcl|lcl}
\hline
\multicolumn{6}{c}{\textbf{Spanish}}\\ \hline	
 \multicolumn{3}{c}{Counselor} & \multicolumn{3}{c}{Client} \\ \hline
 You	&	4.89	&	tu, te, le, usted	&	I	&	4.57	&	yo, conmigo, mi, me	\\
Future	&	3.46	&	enfocaremos, hablaremos, podremos	&	Negate	&	2.29	&	ni, tampoco, nunca, no	\\
We	&	2.34	&	nos, nosotros, nuestra	&	Anger	&	2.06	&	problema, malo, molesta	\\
Achieve	&	1.44	&	dejar, plan, mejorar, controlar	&	Family	&	1.63	&	familiar, padres, hijos	\\
Insight	&	1.27	&	sientes, consideras	&	Negemo	&	1.49	&	enojado, ansiedad, decepcion	\\
Ipron	&	1.21	&	algunos, todos,estas, que	&	Conj	&	1.41	&	pues, y, cuando	\\
Inhib	&	1.16	&	dejar, evitar, control	&	Assent	&	1.41	&	verdad, acuerdo, bien	\\

\hline
\multicolumn{6}{c}{\textbf{English}}\\ \hline		
\multicolumn{3}{c}{Counselor} & \multicolumn{3}{c}{Client} \\ \hline	
You	&	2.04	&	yours, your, you	&	I	&	2.23	&	me, I, myself	\\
We	&	1.59	&	we, us, our	&	Home	&	2.08	&	family, house, room	\\
Cause	&	1.43	&	how, change, control	&	Friend	&	1.67	&	friend, college, partner	\\
Hear	&	1.36	&	sounds, said, hearing	&	Family	&	1.62	&	son, daugher, father, wife	\\
Achieve	&	1.25	&	control, work, able	&	Negate	&	1.46	&	won't, shoudn't, didn't	\\
Percept	&	1.19	&	looking, sound, feel, heard	&	Leisure	&	1.35	&	drinking, playing, exercising	\\
Posemo	&	1.10	&	better, important, fun	&	Discrep	&	1.17	&	if, could, need	\\

\hline\end{tabular}
 \caption{Results from LIWC word class analysis counselor and client interaction in Spanish and English.}
\label{tab:dominance1}

\end{table*}

\noindent \textbf{Language Usage. } We examine language differences using semantic classes from the Linguistic Inquiry and Word Count (LIWC) lexicon \cite{liwc} as a bridge between English and Spanish. The analysis using the Spanish and English LIWC and the word class scoring method of ~\cite{mihalcea2009linguistic} compares the major word categories used by counselors and clients during the conversations. Table~\ref{tab:dominance1} shows the main word classes, with examples,  associated to counselors and clients in both languages. 

Counselors in both languages generally use words related to \textit{you}, \textit{we}, \textit{social}, and \textit{achieve}, which are relevant for MI. However, Spanish MI counselors focus more on \textit{Future} and \textit{Inhib} (inhibition) words. English MI counseling features more \textit{hear} and \textit{percept} (perception) words. These differences could also be related to culture, as in many Spanish-speaking countries healthcare providers take a more authoritative or directive approach to their patients ~\cite{coulter2003european,GimnezMoreno2022TheEO}. In addition,  clients also exhibit similar language use, such as \textit{I}, \textit{Home}, \textit{Family}, \textit{Negate}, with notable differences: Spanish clients use \textit{assent} words, while English clients use \textit{discrep} (discrepancy) words, suggesting greater compliance by Spanish clients.

\begin{figure}[t]
    \centering
\includegraphics[width=.9\columnwidth]{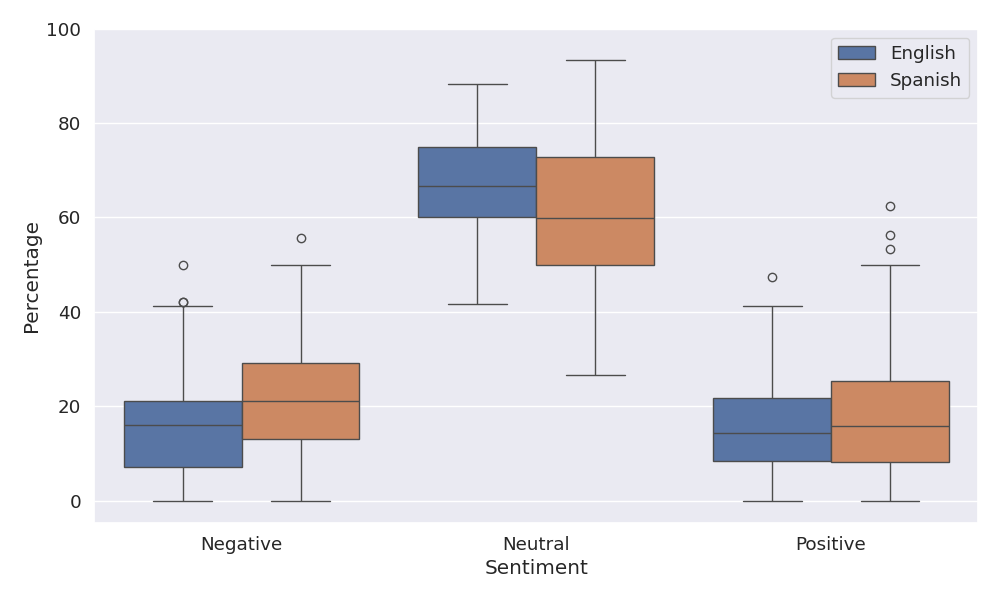} \caption{Counselor sentiment across languages} 
\label{fig:counselor-sentiment-boxplot}
 \end{figure}

\noindent \textbf{Sentiment Trends.}
\label{sec:sentiment} The sentiment exhibited by counselors can reflect their empathy and responsiveness, which are important factors for positive treatment outcome \cite{eberhardt2024decoding,perez-rosas-etal-2019-makes}. 
We use the multilingual PySentimiento library \cite{perez2021pysentimiento} to obtain positive, neutral, and negative sentiment scores on conversational turns. To further evaluate the performance of the sentiment classifier in Spanish data, we randomly sample 10\% (300) of 3,018 Spanish utterances and independently annotate them for sentiment using the same categories. The annotation is conducted by two native Spanish speakers, achieving a Cohen kappa of 0.45 and a raw agreement of 0.64, indicating moderate agreement. A third native speaker conducted further attribution on 107 utterances with disagreement. Among the 300 utterances, the classifier correctly classifies 192, yielding an accuracy of 0.64. Notably, most misclassifications (69 out of 109) occur when the classifier predicts neutral sentiment. Given reasonable accuracy scores, we use classifier predictions to conduct sentiment comparisons across both languages. Figure \ref{fig:counselor-sentiment-boxplot} illustrates the distribution of counselor sentiment, showing that neutral sentiment is the most prevalent in both languages, while positive and negative sentiments occur more frequently in Spanish conversations.

\begin{table*}[t!]
\centering
\small
\begin{tabular}{lll|ll|ll|ll}
\multicolumn{5}{c}{{ \textbf{Monolingual Models}}} & \multicolumn{4}{c}{{ \textbf{Multilingual Models}}} \\ \hline
              & \multicolumn{2}{c}{en-BERT} &        \multicolumn{2}{c}{sp-BETO}       & \multicolumn{2}{c}{en-MLBERT} &        \multicolumn{2}{c}{sp-MLBERT}        \\ \cline{2-9}
              
              & 2-way   & 3-way & 2-way   &   3-way & 2-way     & 3-way & 2-way     & 3-way \\ \cline{1-9} 
Accuracy      & .83   & .88 & .92        & \multicolumn{1}{l|}{.92} & .77     & .89 & .84    & .89 \\ \hline
F1            & .82   & .88 & .92        & \multicolumn{1}{l|}{.92} & .76     & .88 & .84     & .88 \\ \hline
F1-Other       & -       & .95   & -            & \multicolumn{1}{l|}{.90}   & -         & .96   & -         & .95   \\ \hline
F1-Question   & .88     & .65   & .95 & \multicolumn{1}{l|}{.89}   & .84       & .63   & .90       & .54   \\ \hline
F1-Reflection & .64     & .46   & .82 & \multicolumn{1}{l|}{.66}   & .57       & .29   & .68       & .22   \\ \hline
\end{tabular}%
    \caption{ Classification results using monolingual models (sp-BETO, en-BERT) and multilingual models (sp-MLBERT, en-MLBERT) for 2-way (reflection vs question) and 3-way (Question vs Reflection vs Other) classification. Notations in the form \{language-\textsc{model}\} indicate in which language the model is fine-tuned on.}
    \label{tab:3way}
\end{table*}

\section{Predicting Counselor Behaviors}

In addition to linguistic analyzes, we perform classification experiments in Spanish and English conversations to classify counselor behavior using MIDAS and its English counterpart, described in Section 4. Similarly to the label classification experiments in \cite{hebrew-counseling}, we define two tasks: binary classification to differentiate reflections from questions, and three-way classification to identify questions, reflections, or neither. 
We experiment with two settings: we train and test the classifiers using the same language for both the training and the test data; and we use multilingual language models to enable training on one language and evaluation on the other.

For our experiments, we use a 85\%--15\% training--test split. 
For the monolingual experiments, as our main models we use BERT  \cite{devlin2018bert} and BETO \cite{cañete2023spanish}, a BERT architecture trained on Spanish text.
For the multilingual experiments, we use a BERT architecture trained for multiple languages, including English and Spanish BERT \cite{DBLP:journals/corr/abs-1810-04805}, denoted as ML-BERT. We attach classification heads to the base models and fine-tune each model for five epochs each. Results for the classification experiments
are shown in Table \ref{tab:3way}.

In general, we observe that questions are easier to predict than reflections. This aligns with previous work done on English, where reflections were also more challenging to classify, and with work conducted on Hebrew \cite{hebrew-counseling} in which questions are easier to classify than other codes. 
An important take-away from our experiments is that performing training and evaluation  in the same language outperforms multilingual settings.

\section{Conclusion}

In this work, 
we introduced MIDAS, a Motivational Interviewing Dataset in Spanish, the first Spanish MI dataset. We conducted comparative analyzes of the language used by counselors in Spanish and English counseling interactions and found differences in linguistic styles and conversation dynamics.  
Future work includes a more extensive analysis of the differences between English and Spanish counseling, including conversational dynamics such as verbal mirroring and power dynamics, as well as conversational strategies such as empathy or partnership. We also envision MIDAS as a valuable resource in building NLP applications to support counseling evaluation and training for Spanish speakers.



The MIDAS dataset is publicly available under \url{https://github.com/MichiganNLP/MIDAS}.


\section{Limitations} 

A limitation of this work is that the collected transcripts are sourced from online videos created for educational purposes and may be scripted to some extent. However, it is important to mention that in real counseling this is a common practice, as counseling training often makes use of actors who perform different learning scenarios. Although client behavior may be more unpredictable in real counseling, we believe that this dataset can provide important information for the study of the behavioral and cultural differences of Spanish counseling.



\section{Acknowledgments}
We are grateful to Marlene Reyes, Hector Pizarro, and Ana Ronquillo for assisting us with the data collection and the counseling annotations. We also thank the anonymous reviewers for their constructive feedback and the members of the Language and Information Technologies lab at the University of Michigan for the insightful discussions during the early stages of the project. This project was partially funded by a National Science Foundation award (\#2306372). Any opinions, findings, conclusions, or recommendations expressed in this material are those of the authors and do not necessarily reflect the views of the National Science Foundation.

\bibliography{anthology,custom}

\end{sloppypar}
\end{document}